\documentclass{article}

\usepackage{hyperref}
\usepackage{fullpage}
\usepackage{crop}%
\usepackage{amsmath,amssymb,amsfonts,upref,endnotes}
\usepackage{amsthm}
\usepackage{marginnote}
\RequirePackage{graphicx}
\usepackage[figuresright]{rotating}
\usepackage{color}

\begin{document}

\title{Gaussian Process Manifold Interpolation for Probabilistic Atrial Activation Maps and Uncertain Conduction Velocity}

\author{
Sam~Coveney$^{1}$, Cesare~Corrado$^{2}$, Caroline~H~Roney$^{2}$,\\
Daniel~O'Hare$^{2}$, Steven~E~Williams$^{2}$, Mark~D O'Neill$^{2}$,\\
Steven~A~Niederer$^{2}$, Richard~H~Clayton$^{1}$, Jeremy~E~Oakley$^{3}$, and Richard~D~Wilkinson$^{3}$}

\maketitle

\begin{abstract}
In patients with atrial fibrillation, local activation time (LAT) maps are routinely used for characterising patient pathophysiology. The gradient of LAT maps can be used to calculate conduction velocity (CV), which directly relates to material conductivity and may provide an important measure of atrial substrate properties. Including uncertainty in CV calculations would help with interpreting the reliability of these measurements. Here, we build upon a recent insight into reduced-rank Gaussian processes (GP) to perform probabilistic interpolation of uncertain LAT directly on human atrial manifolds. Our Gaussian Process Manifold Interpolation (GPMI) method accounts for the topology of the atria, and allows for calculation of statistics for predicted CV. We demonstrate our method on two clinical cases, and perform validation against a simulated ground truth. CV uncertainty depends on data density, wave propagation direction, and CV magnitude. GPMI is suitable for probabilistic interpolation of other uncertain quantities on non-Euclidean manifolds.
\end{abstract}

\begin{minipage}{1.0\linewidth}
\textbf{Address}\par
$^{1}$Insigneo Institute for in-silico medicine and Department of Computer Science, University of Sheffield, United Kingdom.\\
$^{2}$Division of Imaging Sciences and Biomedical Engineering, King's College London, United Kingdom\\
$^{3}$School of Mathematics and Statistics, University of Sheffield, United Kingdom \\
Corresponding Author: Sam Coveney (s.coveney@sheffield.ac.uk) \\
Subject: Cardiac electrophysiology, Statistical methods, Uncertainty quantification 
\end{minipage}

\section{Background}

Electrical activation of cardiac tissue acts to initiate and synchronise mechanical contraction. The spread of the activation wave is characterised by local activation times (LAT) and conduction velocity (CV), where CV describes the local speed and direction of the wavefront \cite{Cantwell2015}. CV is a clinically important material property of cardiac tissue \cite{Honarbakhsh2019, Bellmann2019}, with decreases in CV often associated with abnormalities such as tissue fibrosis. CV can also be used to calibrate patient-specific models of electrophysiology \cite{Corrado2018}, which have the potential to be used for planning clinical interventions \cite{Boyle2019}. CV can be derived from measurements of LAT at a set of locations. In the clinical setting, LAT can be obtained from electrograms recorded using catheters placed inside the heart. However, reliable CV estimation based on uncertain estimates of LAT from noisy electrograms is difficult \cite{Cantwell2015,Verma2018}. Wave collision and lines of block also present challenges for CV estimation \cite{Loewe2019,Roney2019}.


In a previous study of electrical activation in the human left atrium, we identified uncertainties associated with identifying LAT from individual electrograms, as well as uncertainty associated with the registration of the recording electrode to an anatomical mesh \cite{Coveney2019}.
We demonstrated that it is possible to create probabilistic LAT maps over the left atrium, which is modelled as a 2D manifold in 3D space represented as a triangle mesh. These LAT maps take account of uncertainties in both LAT measurements and the interpolation procedure itself. Importantly, the interpolation was performed directly on the manifold geometry, taking account of the physical distances travelled by the wavefront and the connectivity between points on the mesh. We used a stochastic partial differential equations (SPDE) approach linking  Gaussian Markov random fields (GMRF) to a class of Gaussian processes (GP) with Mat\'ern covariance functions \cite{Lindgren2011}. This enabled us to construct random fields on non-Euclidean domains such as the manifold describing the left atrium.

Given a probabilistic LAT map on an atrial mesh, a CV map can easily be calculated from the LAT map posterior mean. A natural approach to compute how our uncertainty about LAT affects our estimate of CV is to use stochastic sampling. We can simulate samples from our posterior distribution for LAT, calculating CV for each of them, thus estimating a distribution for CV at every mesh location. However, doing this with the GMRF model used in \cite{Coveney2019} shows that this model is not suitable for estimating uncertainty about CV, because random samples of LAT are insufficiently smooth. In practical terms this leads to  samples of CV that are not consistent with observations of LAT. In technical terms, the Mat\'ern class of covariance functions depends upon a smoothness parameter $\nu$, such that process samples are $k$-times mean-squared differentiable if and only if $\nu > k$ \cite{Rasmussen2006}. For the  SPDE-GMRF, $\nu$ is constrained to be one, so that posterior LAT samples are not differentiable. Consequently, they are too rough to represent a spreading electrical wavefront, meaning the approach cannot be use to calculate the CV distribution. In this paper, we describe a new GP approach that does allow us to estimate CV distributions from a probabilistic interpolation of noisy LAT measurements.

The reason for using the SPDE-GMRF approach was that an `ordinary GP', for which the covariance matrix is a function of the Euclidean distance matrix between 3D spatial locations, would not account for the geometry or topology of the atrial manifold. It is not possible to simply replace  Euclidean  distances with for example geodesic distances because the resulting covariance matrix is not positive semi-definite. However, the meaningful distance between locations for this physical system is geodesic distance on the manifold, not distance through the manifold that ignores the topology. A recent advance in reduced-rank Gaussian processes showed that it is possible to expand a stationary covariance kernel in terms of eigenfunctions of the Laplacian operator on the spatial domain  \cite{Solin2019}.  The contribution of this paper is to address the problem of obtaining distributions of CV from probabilistic interpolation of LAT. We apply this recent advance to a 2D manifold in 3D space, so we are able to form a Gaussian process on the manifold using any stationary covariance function, including those which lead to differentiable samples. This allows us to calculate CV distributions by probabilistic interpolation of LAT.

This paper is organised as follows. We first explain how to  interpolate LAT using ordinary Gaussian processes. We then review how to expand the covariance function as in \cite{Solin2019}, but applied to a 2D triangulated manifold representing the atrium, such that we can simply replace the covariance matrix in ordinary GP regression to obtain a GP confined to the manifold. Using clinical data, we perform LAT interpolation and obtain corresponding probabilistic CV maps. We also validate our methodology against a simulated ground truth. We find that CV uncertainty depends on LAT observation uncertainty, data density, alignment of observations with respect to wave propagation direction, and on CV magnitude. We then discuss future work made possible by solving the geometrical aspect of probabilistic interpolation on atrial manifolds.


\section{Methods}\label{sec:methods}

\subsection{Data collection}

As proof of concept, we present two clinical cases in this paper. Additional details on data collection are given in the Supplementary Material. For the first case, Patient 1, (identical to Patient A in \cite{Coveney2019}) observations were collected using an S1-S2 pacing interval: a 2-beat drive train with a cycle length of S1 $470$ms followed by a single premature extra stimulus S2, where S2 < S1. For the second case, Patient 2, a constant pacing at cycle length $500$ms was used, i.e., S1 only. For Patient 1, electrograms were processed to obtain uncertain LAT as described in \cite{Coveney2019}. For Patient 2, visual inspection of each point was undertaken by one of the authors to assign LAT to the earliest sharp deflection, without an assigned uncertainty based on the electrogram analysis.

In both cases, a triangulated mesh of the left atrium, obtained using the respective mapping systems, was processed to clip away the pulmonary veins and mitral valves to leave holes, and the triangulation was smoothed (using mainly surface-preserving Laplacian smoothing, element sub-division, and short-edge collapsing) to be reasonably regular, with $\sim 10^4$ non-intersecting mesh faces. This yields an interpolation resolution mesh, in which LAT observations were assigned to the nearest vertex, and given an additional observation uncertainty arising from this mesh assignment as in \cite{Coveney2019}. Note that our method does not require observation-specific uncertainty for LAT, or specific methods for determining LAT and assigning observations to a mesh.

\subsection{Gaussian process regression}

A function $f$, for which only noisy observations $y_i$ are available at spatial locations $\mathbf{x}_i$, can be modelled by a Gaussian process as follows:
\begin{align}
    f(\mathbf{x}) \sim \mathcal{GP}(0, \tau^2 k(\mathbf{x}, \mathbf{x'})), \notag \\
    y_i = f(\mathbf{x_i}) + \epsilon_i,   \quad \text{where} \;\;  \epsilon_i \sim \mathcal{N}(0, \sigma_i^2), \label{eq:GP}
\end{align} 
where $\epsilon_i$ is the (heteroscedastic) noise for each observation. The covariance function or "kernel" $k$ specifies the covariance $\text{cov}(f(\mathbf{x}), f(\mathbf{x}')) := \tau^2 k(\mathbf{x}, \mathbf{x}')$ as a function of location. Here, the mean of the Gaussian process is set to zero. We can generalize by using a non-zero parametric mean function, but by centering and scaling the data $y$ to be mean zero, variance one, we find that a zero mean is sufficient and simplifies the exposition. A GP is 'fit' to data by conditioning upon the observations $\mathcal{D}=\{\mathbf{x_i}, y_i\}_{i=1}^n$ and optimizing $\tau^2$ and any kernel hyperparameters to maximize the log-likelihood. Here $\sigma_i$ is obtained by pre-processing LAT and mesh assignment rather than by optimization, although such optimization would be possible.

Posterior predictions (i.e., using the GP conditioned upon the data) are multivariate Gaussian distributions specified by the posterior mean  and variance. At input (e.g., location) $\mathbf{x^*}$, the posterior mean and variance is given by:
\begin{align}
    \begin{split}
    \mathbb{E}[f(\mathbf{x^*}) \mid \mathcal{D}] &= \mathbf{k_*^T} (\mathbf{K} + \boldsymbol{\Sigma} )^{-1} \mathbf{y}  \\
    \mathbb{V}[f(\mathbf{x^*})\mid \mathcal{D}] &= \tau^2 k(\mathbf{x^*}, \mathbf{x^*}) - \mathbf{k_*^T} (\mathbf{K} + \boldsymbol{\Sigma})^{-1} \mathbf{k_*}
    \end{split}\label{eq:GPpred}
\end{align}
where $\mathbf{K}$ is an $n\times n$ matrix containing the covariance between all pairs of observations $\tau^2 k(\mathbf{x_i}, \mathbf{x_j})$; $\mathbf{y}$  is the vector of observations $y_i$; and $\boldsymbol{\Sigma} \equiv \text{diag}(\sigma_1^2 + \eta, \dotsc , \sigma_n^2 + \eta)$ contains the observation error variances $\sigma_i^2$ (for homoscedastic noise, all $\sigma_i$ are equal) plus a `nugget' $\eta$ representing additional observation uncertainty common to all observations ($\eta$ is optimized with other hyperparameters). The cross-covariance between observations and predictions is given by the vector $\mathbf{k_*}$, and $\tau^2 k(\mathbf{x^*}, \mathbf{x^*})$ is the covariance between prediction locations (both are matrices when $\mathbf{x}^*$ contains multiple locations). By predicting at multiple locations simultaneously, e.g., all points on the atrial mesh, we obtain expressions for the posterior mean and variance at all locations.

%

\subsection{LAT interpolation with a Gaussian process}
\label{sec:LATwithGP}

We assume the underlying LAT field is a smoothly varying function of location and model it as a Gaussian process: $\text{LAT} \equiv f(\mathbf{x}) \sim \mathcal{GP}(0, \tau^2 k(\mathbf{x}, \mathbf{x'}))$. We model the relationship between the noisy LAT \emph{observations} $\mathbf{y}$ and the `true' LAT as $y_i = f(\mathbf{x_i}) + \epsilon_i$ where $\epsilon_i \sim \mathcal{N}(0, \sigma_i^2)$, as in Eq \eqref{eq:GP}. LAT observation $y_i$ is assigned to a coordinate on the atrial mesh $\mathbf{x_i}$.


For the GP kernel, we choose the Mat\'ern 3/2 kernel ($\nu = 3/2)$, which gives a once (mean squared) differentiable process. This choice was made because CV is obtained from the derivative of LAT, but we do not want to make assumptions about the smoothness of CV. The Mat\'ern 3/2 kernel is a stationary kernel that, when working in Euclidean space, depends only on the Euclidean distance $r_{ij}=||\mathbf{x}_i-\mathbf{x}_j||_2$ between observations $i$ and $j$, i.e. $k(\mathbf{x}_i, \mathbf{x}_j) \equiv k(r_{ij})$, and can be written, with length-scale hyperparameter $l$, as:
\begin{equation} \label{eq:matern52}
   k_{3/2}(r) = \left(1 + \frac{\sqrt{3}r}{l} \right) \exp{ \left( - \frac{\sqrt{3}r}{l} \right) }.
\end{equation}
%

In \cite{Coveney2019}, we jointly interpolated LAT across both space and S1-S2 pacing using an order one auto-regressive model (i.e., AR(1)) for the dependence on S2. We do not interpolate over S2 in this paper,
but could do so using a separable covariance structure such as  $ k_{\text{spatial-pacing}}((x,S2), (x', S2')) = k_{3/2}(x,x') \times k_\text{pacing}(S2, S2')$, where $k_{\text{pacing}}(S2, S2')$ is a suitable standard kernel (e.g. squared exponential), simplifying  the approach from \cite{Coveney2019}.

\subsection{Gaussian processes on a manifold}



The standard GP approach described above assumes, due to the kernel, that correlations are a function of Euclidean distance. Since electrical activation advances over the atrial manifold, instead of working with Euclidean distances $||\mathbf{x} - \mathbf{x}'||_2$ in the domain $\mathbb{R}^3$, we want to limit the domain to locations on the atria, which we denote by $\Omega$, and assume correlations are a function of geodesic path length on that domain. Unfortunately, it is not possible to simply replace the Euclidean distance in a kernel such as Eq. \eqref{eq:matern52} with the geodesic distance and still form a valid covariance matrix, as this leads to non-positive definite kernels \cite{Wood1995}.

Here, we take advantage of a recently developed \cite{Solin2019, Kok} reduced-rank method for expressing a stationary kernel in terms of the kernel spectral density and eigensolutions of the Laplacian operator on $\Omega$, that allows us to bypass the need to explicitly define a covariance function. We choose a spectral density, and then define  a stochastic process with that spectral density, which then implicitly defines a covariance function on $\Omega$ (by the Wiener-Khinchin theorem).

We use a finite dimensional model of the form
\begin{equation} 
\tilde{f}(\mathbf{x}) = \sum_{k=1}^M f_k \phi_k(\mathbf{x}) \quad \mbox{ where } \quad f_k \sim N(0, S(\sqrt{\lambda_k})), \label{eqn:expansion}
\end{equation}
where the $f_k$ are parameters to be learnt.  
$\tilde{f}(x)$ is a (finite dimensional) Gaussian process with covariance function given by the basis function expansion
\begin{align}\label{eq:kernel}
   \mathbb{C}\operatorname{ov}(\tilde{f}(\mathbf{x_i}), \tilde{f}(\mathbf{x_j}))= \tilde{k}(\mathbf{x_i},\mathbf{x_j}) = \sum_{k=1}^{M} S(\sqrt{\lambda_k}) \phi_k(\mathbf{x_i}) \phi_k(\mathbf{x_j}). 
\end{align}
The basis vectors $\phi_k(\mathbf{x})$ are chosen to be eigenfunctions corresponding to eigenvalues $\lambda_k$ of the negative Laplacian operator on the domain $\Omega$, i.e., $\{\lambda_k, \phi_k(\cdot)\}$ are solutions of the system 
\begin{align}
\begin{split}
    - \nabla^2 \phi_k(\mathbf{x}) = \lambda_k \phi_k(\mathbf{x}), \quad{} &\mathbf{x} \in \Omega,  \\
    \nabla \phi_k(\mathbf{x}) \cdot \vec{n}_{\partial\Omega} = 0, \quad \quad &\mathbf{x} \in \partial\Omega
\end{split} \label{eq:eigenLaplacian}
\end{align}
with $0\leq\lambda_1 < \lambda_2 < \ldots$. Note that this basis  is independent of our choice of covariance structure/spectral density, allowing us to try different structures without needing to recompute the basis vectors.
We use Neumann boundary conditions specifying zero derivative normal to the mesh edges (pulmonary veins and mitral valve), as these do not specify LAT in advance of the interpolation (in contrast \cite{Solin2019, Kok} used Dirichlet boundary conditions).

$S(w)$ is a spectral density \cite{Rasmussen2006} which we are free to choose;  it is  $S$ that determines the nature of the covariance structure of our model. In this paper,  we use the spectral density corresponding to the  Mat\'ern family of covariance functions, which is given by
%
\begin{align}
    S(w) := \;\; \frac{2^D \pi^{D/2} \Gamma(\nu + D/2) (2 \nu)^{\nu}  }{ \Gamma(\nu) l^{2\nu} }  \left( \frac{2\nu}{l^2} + 4 \pi^2 \omega^2  \right)^{-(\nu + D/2)}\label{eq:SpectralDensity}
\end{align}
where $\nu$ is smoothness, $l$ is length-scale, $D$ is dimensionality ($D = 2$ for a 2D manifold in 3D space), and $\Gamma$ is the Gamma function. 
The spectral density of our model $\tilde{f}$ is only approximately $S(w)$, but converges to $S(w)$ as $M$ and $\Omega$ grow \cite{Solin2019}.

The motivation in \cite{Solin2019} for using expansion (\ref{eqn:expansion}) is to reduce the computational complexity of hyperparameter estimation, as the basis is independent of the covariance function. In contrast, our motivation for using this approach is that the solution of Eq. \eqref{eq:eigenLaplacian} can be computed numerically on a discrete mesh, even for non-Euclidean domains, allowing us to define a GP model directly on the manifold defined by $\Omega$. In particular, we have chosen to model LAT on an atrial manifold using a stochastic process with Mat\'ern spectral density. 


\subsection{Eigenfunction calculations} \label{sec:eigenpairs}

\begin{figure}[h]
	\centering
    \caption{Left: representation of the terms in equation \eqref{eq:laplacian}, where the shaded area represents the Voronoi cell with area $A_i$ and the angles opposing the edge between vertex $i$ and $j$ are $a_{ij}$ and $b_{ij}$. Right: subdivision of a mesh triangle face into 9 triangles by addition of vertices, including a vertex at the mesh face centroid.} \label{fig:mesh-stuff}
    \includegraphics[width=0.275\textwidth]{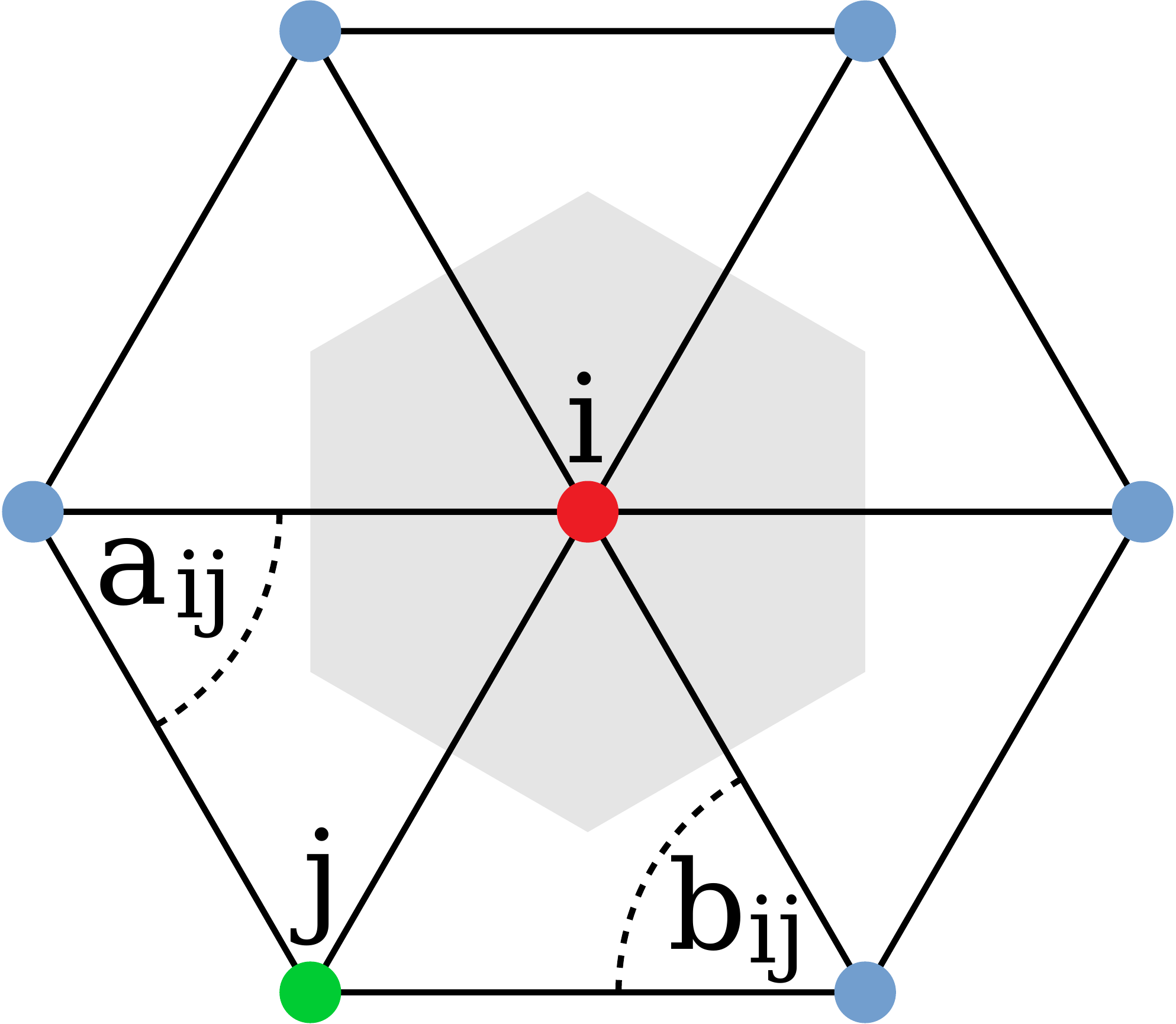}
	\hspace{2.0cm}
    \includegraphics[width=0.275\textwidth]{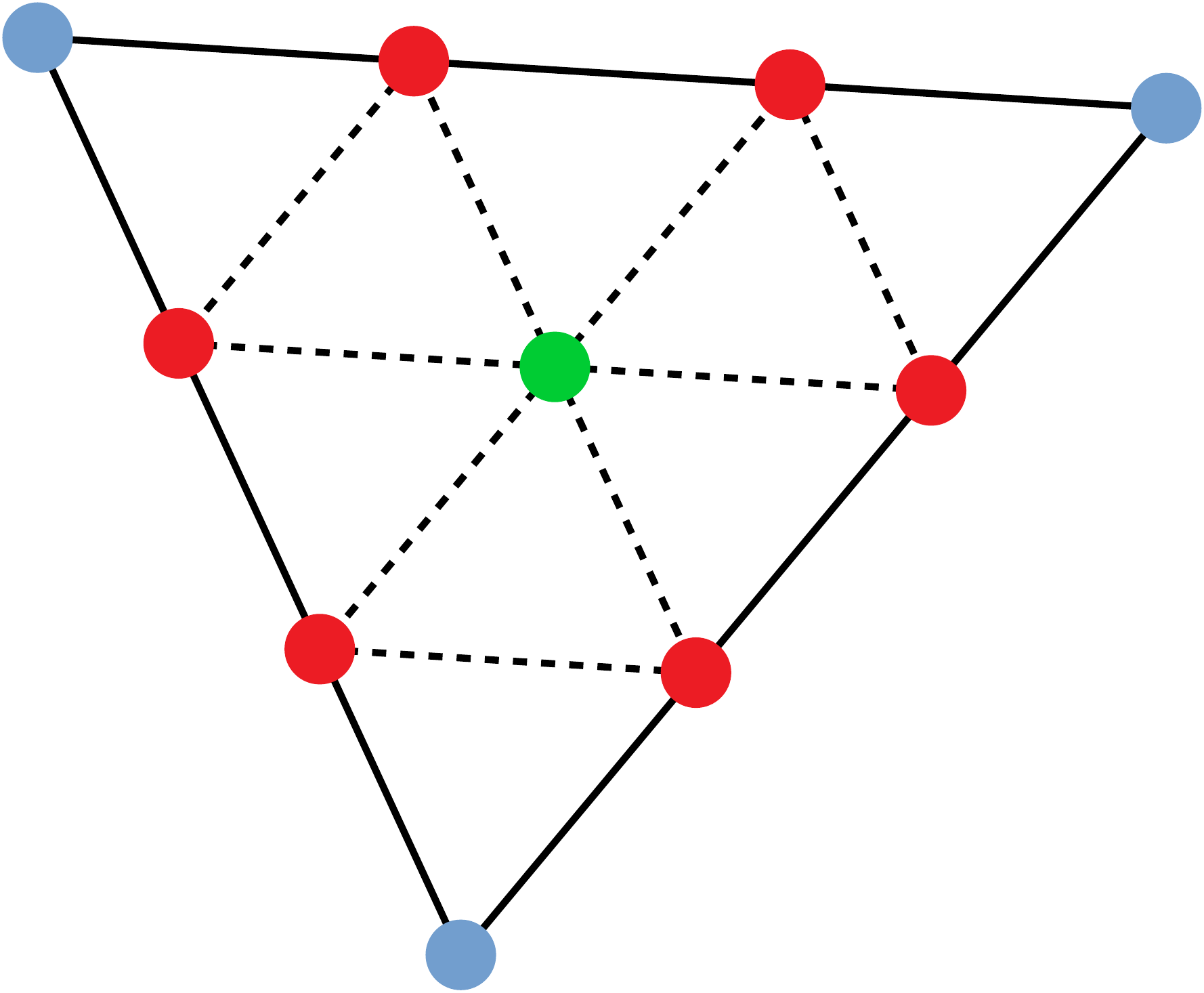}
\end{figure}

To apply this kernel expansion \eqref{eq:kernel} to a 2D manifold in 3D space, we represent the manifold as a discrete triangle mesh with $N$ vertices and represent the Laplacian operator with the 'mesh/cotan Laplacian' (see \cite{Sorkine2005}) given by:
\begin{equation}\label{eq:laplacian}
    (\nabla^2 s)_i = \frac{1}{2A_i} \sum_{j \in N(i)} (\cot{a_{ij}} + \cot{b_{ij}})(s_i - s_j)
\end{equation}
where $s$ represents an arbitrary scalar field defined at vertices on the mesh. Figure \ref{fig:mesh-stuff} (left) represents the terms in eq. \eqref{eq:laplacian}. Applied to a triangular mesh, the $N \times N$ operator matrix $\mathcal{L}$ represents the Laplacian with Neumann boundary conditions at the mesh edges (mitral valve and pulmonary veins). We calculate the Laplacian matrix $\mathcal{L}$ neglecting the areas $A_i$, and solve the eigenvalue problem $\mathcal{L} \phi = \lambda A \phi$ where $A$ is a diagonal matrix of areas $A_i$ (see \cite{Reuter2009}), solving for the M smallest eigenvalues and corresponding eigenvectors using the \emph{eigsh} function from the Python package \emph{Scipy} \cite{SciPy}.
%

We obtain $M$ eigenvectors such that $\phi_k(\mathbf{x_i}) \equiv \phi_k^i$, where $k = 1 \dotsc M$ index the eigenfunction and $i = 1 \dotsc N$ index the mesh vertices. These eigenfunctions belong to the spatial domain represented by the mesh, and as such the topology (connectivity) and geometry (distances) of the atrial manifold are properly accounted for. Note that the eigenproblem only needs to be solved once for a particular mesh. We can then construct all of the covariance matrices needed for GP regression. We refer to this method as Gaussian Process Manifold Interpolation (GPMI). Our code for applying GPMI to atrial manifolds can be found at \cite{quLATi}.

To represent the atrial mesh smoothly, we use an interpolation resolution mesh of $\sim 10^4$ vertices which corresponds to an edge length $\sim 1$mm. In order to ensure a fine discretisation of the Laplacian, and also to allow the calculation of eigenfunction gradients (required for the posterior of the GP gradient; see Section \ref{sec:methods}\ref{sec:CV}), we subdivide the interpolation mesh: each triangular element (3 vertices) is subdivided into 9 elements by creating two additional vertices along each edge and one additional vertex at the element centroid, as shown in figure \ref{fig:mesh-stuff} (right). The Laplacian and the corresponding eigenfunctions are calculated for this finer mesh.

For each original mesh face, there are now 10 vertices which lie in the plane of this face. To calculate the gradients of the eigenfunctions at a face centroid, we first rotate these (finer mesh) vertex coordinates into a 2D frame, and then fit a cubic polynomial to the vertex eigenfunction values and 2D coordinates using least squares regression. We then evaluate the gradient of the eigenfunctions at the centroid using the derivative of the polynomial expression, and rotate this (vector) gradient back into the original 3D coordinate frame. This allows us to calculate the eigenfunction gradients at the centroids, even when the eigenfunctions vary non-linearly across a face. For simplicity, we retain the eigenfunction values at the mesh vertices and face centroids of the original mesh only (i.e. we discard the vertices from subdividing the face edges), which allows us to assign LAT observations and make LAT predictions at both vertices and centroids.

Boundary conditions at the edges of the mesh (holes representing the mitral valve and pulmonary veins) will leave unwanted artefacts in the interpolation. To deal with this problem, after forming the interpolation resolution mesh and prior to subdivision as described above, we append tubes of triangular elements to the atrial holes, consisting of 10-20 layers of elements (depending on the mesh resolution) and extending away from the mesh centre of mass, using a custom algorithm described in the Supplementary Material. This takes the boundaries away from the 'real atrium' and removes the unwanted boundary effects on the interpolation. Note that the topology is unchanged by this procedure. After subdivision as described above, we retain only the eigenvector entries corresponding to the original mesh vertices and centroids before extension. A figure of the mesh extension result for Patient 2 is given in the Supplementary Material.

\subsection{Conduction Velocity Distribution} \label{sec:CV}

Conduction velocity is the inverse of the LAT gradient. The distribution for either the LAT gradient or CV could be calculated by first generating samples from the posterior distribution for LAT, and then calculating the desired quantities for each sample. However, since applying a linear operator (such as the gradient) to a Gaussian process results in another Gaussian process, the gradient of LAT is also a Gaussian process. Therefore, we can instead sample directly from this gradient distribution. The posterior distribution for the gradient of LAT is given by similar expressions to the posterior distribution for LAT (eq. \ref{eq:GPpred}), except that the covariance entries involving prediction locations require derivatives of the covariance function with respect to these prediction locations:
\begin{align}
    \begin{split}
    \mathbb{E}\left[\frac{\partial f(\mathbf{x^*})}{\partial \mathbf{x^*}} \mid \mathcal{D}\right] &= \frac{\partial \mathbf{k_*^T}}{\partial \mathbf{x^*}} (\mathbf{K} + \boldsymbol{\Sigma} )^{-1} \mathbf{y}  \\
    \mathbb{V}\left[\frac{\partial f(\mathbf{x^*})}{\partial \mathbf{x^*}} \mid \mathcal{D}\right] &= \tau^2 \left.\frac{\partial ^2 k(\mathbf{x_a}, \mathbf{x_b})}{\partial \mathbf{x_a} \partial \mathbf{x_b}}\right|_{\mathbf{x_a} = \mathbf{x_b} = \mathbf{x^*}}  - \frac{\partial \mathbf{k_*^T}}{\partial \mathbf{x^*}} (\mathbf{K} + \boldsymbol{\Sigma})^{-1} \frac{\partial \mathbf{k_*}}{\partial \mathbf{x^*}}
    %
    %
    \end{split}\label{eq:GPpred_gradient}
\end{align}
When the kernel is expressed as in equation \eqref{eq:kernel}, the necessary kernel derivatives are given as:
\begin{align}
\begin{split}
    \frac{\partial k(\mathbf{x_i}, \mathbf{x_j})}{\partial \mathbf{x_i}} &=  \sum_{k=1}^{M} S(\sqrt{\lambda_k}) \frac{\partial \phi_k(\mathbf{x_i})}{\partial \mathbf{x_i}} \phi_k(\mathbf{x_j}) \\
    \frac{\partial^2 k(\mathbf{x_i}, \mathbf{x_j})}{\partial \mathbf{x_i} \partial \mathbf{x_j}}  &=  \sum_{k=1}^{M} S(\sqrt{\lambda_k}) \frac{\partial \phi_k(\mathbf{x_i})}{\partial \mathbf{x_i}} \frac{\partial \phi_k(\mathbf{x_j})}{\partial \mathbf{x_j}}. 
\end{split}\label{eq:kernel_gradients}
\end{align}
The derivatives of the eigenfunctions can all be calculated in advance of the interpolation, after calculating the eigenfunctions themselves (see Section \ref{sec:eigenpairs}). Note that it is therefore possible to include uncertain \emph{observations} of the gradient of LAT into the modelling.

Given the gradients of the eigenfunctions at the face centroids (see Section \ref{sec:eigenpairs}), the posterior distribution of the gradient of LAT can be calculated at all centroids.
However, since neither magnitude or inverse are linear operations, we calculate the distributions for the magnitude of the gradient of LAT and CV by sampling from the posterior LAT gradient and applying the operations to the samples. At each centroid, we generate $2000$ samples from the multivariate normal distribution for the components of the gradient of LAT given by equation \eqref{eq:GPpred_gradient}. We calculate the standard deviation, 9th, 25th, 50th, 75th, 91st percentiles for the magnitude of the posterior samples of the gradient of LAT (the percentiles can be inverted to obtain CV percentiles). We use the posterior mean of \eqref{eq:GPpred_gradient} to represent prediction of the magnitude, and statistics from sampling for prediction uncertainty.

\subsection{Simulations}

To quantitatively evaluate our methodology, we compare predictions of LAT and CV against a simulation `ground truth'. A computational mesh with a regular triangulation with edge length of $215\mu$m was obtained for Patient 1 ($1047446$ vertices). Simulations were performed with the Cardiac Arrhythmia Research Package (CARP), an electrophysiology solver suitable for high-performance computing \cite{Vigmond2003}. We simulate S1 pacing from the coronary sinus for S1 $750$ms, using the modified Mitchell-Schaeffer ionic model with heterogeneous parameters to generate a spatially heterogeneous conduction velocity field (details in Supplementary Material). To create a `virtual patient' from these simulations, we downsample the simulation mesh to an interpolation resolution mesh, and match the interpolation mesh vertices to the nearest simulation mesh vertices. For each interpolation mesh vertex, we record LAT from the matching simulation vertex.

We calculate CV for the simulation in two ways. Firstly, we calculate CV element-wise at each interpolation centroid from LAT at the face vertices ('element' method): given a face defined by anti-clockwise vertices $\left\{i,j,k\right\}$, face area $A$, face normal $\vec{N}$, edge vectors $\vec{e}_{ab}$, with LAT at vertex $l$ given by $t_l$, the centroid LAT gradient and CV are given by ($\times$ denotes the vector product)
\begin{equation}
    \centering
    \text{CV} = \left. v \middle/ \left| v \right|^2 \right. \quad \text{where} \quad
    v \equiv \nabla \text{LAT} = - \frac{1}{2A}  \vec{N}  \times  \left( t_i \vec{e}_{jk} + t_j \vec{e}_{ki} + t_k \vec{e}_{ij} \right)
\end{equation} \label{eq:CVcalc}

Secondly, since the CV field is heterogeneous and CV often varies on a length-scale smaller than the distance between observations, we also calculate CV taking account of a larger area ('wave' method). For the $20$ vertices nearest to where CV is to be calculated, the vertex coordinates are projected into 2D such that the geodesic distances are optimally conserved. A least squares linear fit to LAT is used to calculate a gradient corresponding to plane wave propagation \cite{Roney2019}. We do this for 1000 random centroids locations, optimized to be well spaced over the mesh using a a maximin distance criterion over $10,000$ designs. Note that $20$ vertices corresponds to at least two 'rings' of vertices around a centroid.

To validate our approach for predicting LAT and CV, we calculate the \emph{normalized} root mean square error (nRMSE), reported as a percentage, and independent standard error (ISE) for all predictions $\mathbf{E}[z_i]$ (with variance $\mathbf{V}[z_i]$) versus ground truths $z_i$, where nRMSE ($\%$) and ISE are given by:
\begin{align} \label{eq:validation}
\begin{split}
	\text{nRMSE} &= \frac{100}{ \text{range} } \sqrt{ \frac{1}{N} \sum_{i=1}^N (\mathbf{E}[z_i]-z_i)^2  }
    \\ 
	\text{ISE}_i &= \left. (\mathbf{E}[z_i]-z_i) \middle/ \sqrt{\mathbf{V}[z_i]} \right.
\end{split}
\end{align}
where $\text{range} = \text{max}(z_i) - \text{min}(z_i)$.

For LAT, about $95\%$ of ISE scores should lie within the interval $\pm 2$. We emphasise that LAT, and therefore (implicitly) the gradient of LAT, is what is directly modelled in our framework. Therefore, for comparison with the ground truth, we predict and compare against the magnitude of the gradient of LAT, rather than against CV, using the standard deviation of the magnitudes of LAT gradient samples. This prevents the nRMSE and ISE scores from being heavily distorted by the inversion involved in calculating CV from the LAT gradient (see Section \ref{sec:Discussion}). We calculate these scores for the LAT gradient for comparison with the two methods of calculating CV outlined above. The meaning of the terms in equation \eqref{eq:validation} for the magnitude of the LAT gradient must be explained: we use the magnitude of the posterior mean of the LAT gradient as $\mathbf{E}[z_i]$, and the standard deviation of the magnitude of the posterior LAT gradient samples as $\sqrt{\mathbf{V}[z_i]}$.

\section{Results}\label{sec:Results}

The number of interpolation mesh vertices and centroids, and the number of observations, are given in Table \ref{tab:numbers}. For the virtual patient, the number of observations will be given in the text for each case. We use $M = 256$ basis functions, which we determined to be sufficient for modelling LAT (see Section 3 in the Supplementary Material).

\begin{table}[!htbp]
\caption{Number of interpolation mesh vertices and centroids, and number of observations, for each case.} \label{tab:numbers}
\centering
\begin{tabular}{|l|ccc|}
\cline{1-4}
Case       &  vertices  &  centroids  & observations  \\ \hline
Patient 1  &  9460      &  18680      & 61            \\
Patient 2  &  17235     &  34098      & 1228          \\
Virtual    &  45180     &  89744      & various       \\ \hline
\end{tabular}
\end{table}

\begin{table}[h]
\caption{Validation results for prediction against ground truth (simulation), averaged over $10$ predictions using different random observation data. 'Wave' and 'Element' refer to methods for calculating CV described in Section \ref{sec:CV}, but the corresponding validation scores are for magnitude of LAT gradient. Both nRMSE and ISE coverage are in $\%$.} \label{tab:val}
\centering
\begin{tabular}{|l|cc|cc|cc|}
\cline{1-7} Number of
& \multicolumn{2}{c|}{LAT} & \multicolumn{2}{c|}{Wave $|\nabla\text{LAT}|$} & \multicolumn{2}{c|}{Element $|\nabla\text{LAT}|$} \\
observations &  nRMSE &  $|\text{ISE}| \leq 2$ &  nRMSE  &  $|\text{ISE}| \leq 2$  & nRMSE & $|\text{ISE}| \leq 2$ \\ \hline
50     &  2.34  &  95.7  &  15.0  &  91.8  &  14.0  &  91.5  \\
100     &  1.51  &  94.3  &  11.7  &  93.7  &  11.1  &  93.4  \\
250     &  0.91  &  96.0  &  9.46  &  94.5  &  9.16  &  93.9  \\
500    &  0.66  &   96.1  &  8.38  &  93.1  &  8.22  &  92.2  \\
750     &  0.56  &   96.1  &  7.62  &  93.3  &  7.56  &  92.2  \\
1000   &  0.50  &   95.6  &  7.22  &  92.6  &  7.22  &  91.6  \\ \hline
\end{tabular}
\end{table}

Validation results for the virtual patient are given in Table \ref{tab:val}, for prediction of LAT and the magnitude of LAT gradient. For each table row, LAT observations were chosen by picking $n$ random vertices and corresponding LAT values, and corrupting these values with additive random noise $\epsilon_i \sim \mathcal{N}(0,1)$. The nRMSE and ISE scores were calculated and averaged across 10 different observation designs for each $n$.

\subsection{LAT interpolation}

\begin{figure}[t]
	\centering
    \caption{Posterior LAT distribution for Patient 1, shown as CV vectors at face centroids colored by LAT posterior mean (top) and standard deviation (bottom). Spheres represent LAT observations $y_i$ (top) and observation noise $\sigma_i$ (bottom).} \label{fig:LAT-apsaf}
       \includegraphics[width=0.475\textwidth]{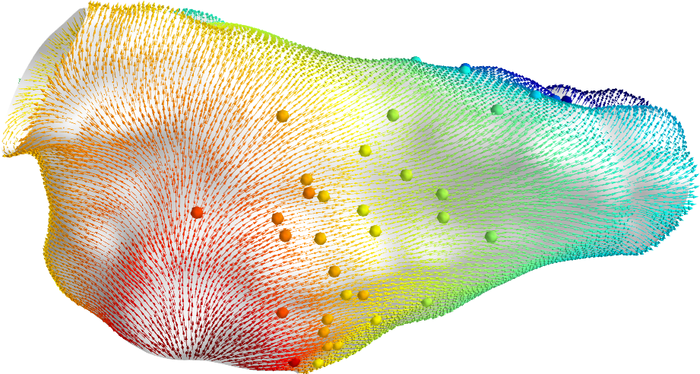}
       \includegraphics[width=0.475\textwidth]{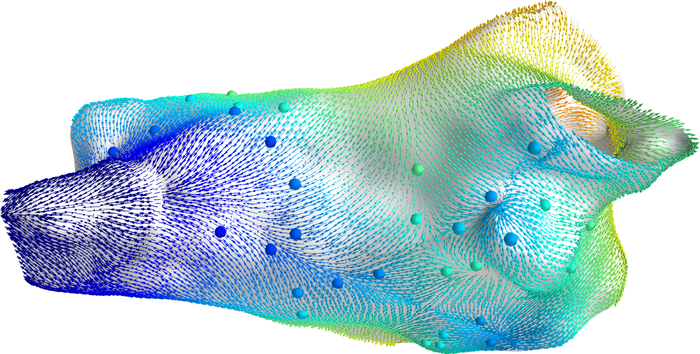}
    \\ \vspace{0.20cm}
       \includegraphics[width=0.5\textwidth]{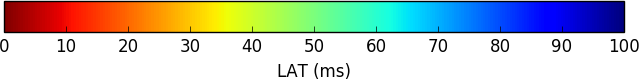}
    \\
       \includegraphics[width=0.475\textwidth]{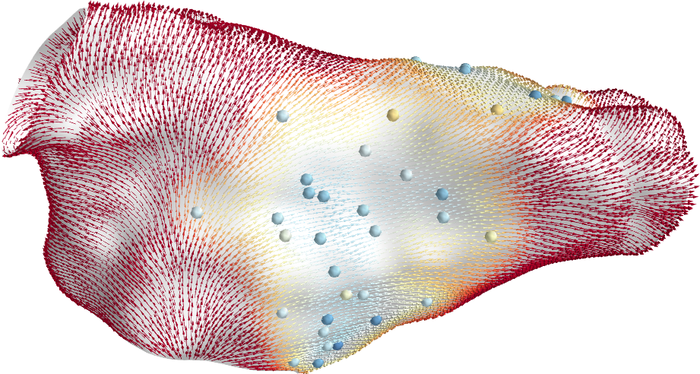}
       \includegraphics[width=0.475\textwidth]{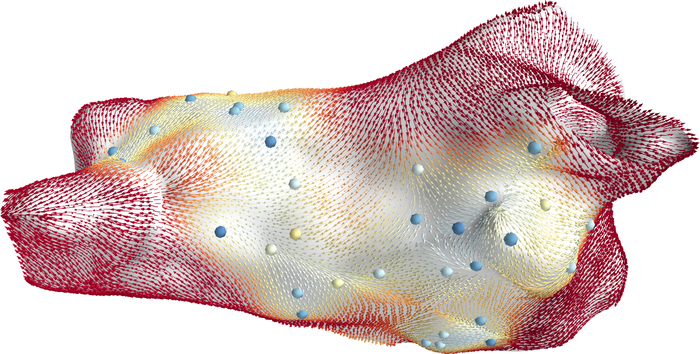}
    \\ \vspace{0.20cm}
       \includegraphics[width=0.5\textwidth]{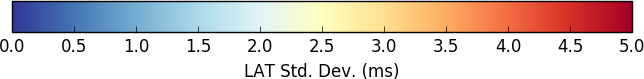}
\end{figure}

\begin{figure}[!htbp]
	\centering
    \caption{Posterior LAT distribution for Patient 2, shown as CV vectors at face centroids colored by LAT posterior mean (top) and standard deviation (bottom). Spheres represent LAT observations $y_i$ (top) and observation noise $\sigma_i$ (bottom).} \label{fig:LAT-S1}
       \includegraphics[width=0.450\textwidth]{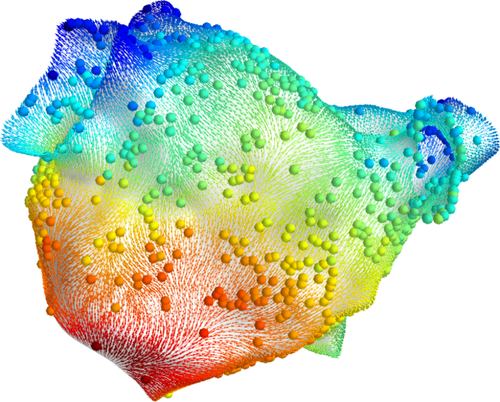}
       \includegraphics[width=0.4250\textwidth]{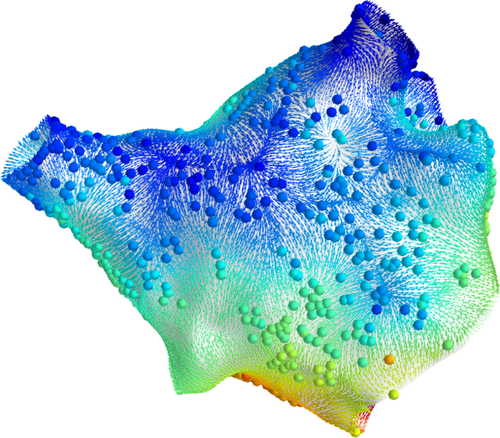}
    \\ \vspace{0.20cm}
       \includegraphics[width=0.50\textwidth]{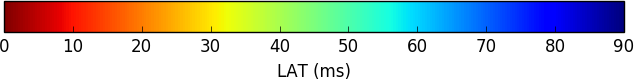}
    \\
       \includegraphics[width=0.450\textwidth]{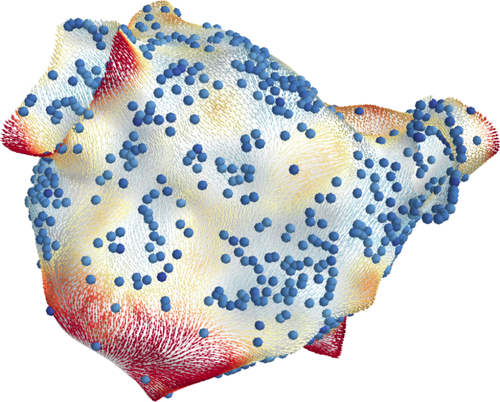}
       \includegraphics[width=0.4250\textwidth]{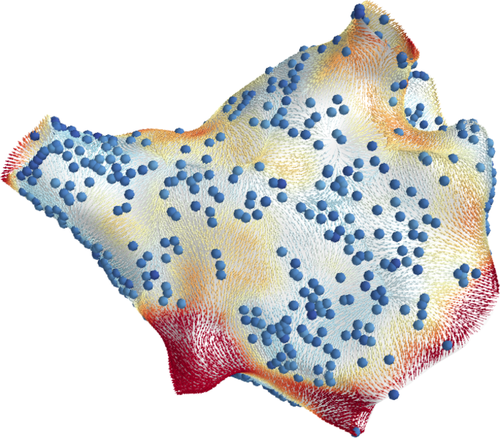}
    \\ \vspace{0.20cm}
       \includegraphics[width=0.50\textwidth]{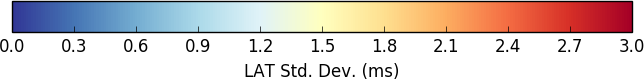}
\end{figure}

Figure \ref{fig:LAT-apsaf} shows LAT interpolation for Patient 1, overlaid with CV vectors coloured by the LAT posterior mean (top) and standard deviation (bottom) at the face centroids. The LAT observations $y_i$ and noise standard deviations $\sigma_i$ are shown as colored spheres. 
LAT prediction uncertainty depends on the distance from the observations and on the observation uncertainty. These interpolation results are comparable with those in \cite{Coveney2019}.
Figure \ref{fig:LAT-S1} shows the LAT interpolation for Patient 2. In this case, the observation noise only arises from assigning observations to mesh vertices and not from LAT assignment to electrograms, resulting in a very small observation noise. Prediction uncertainty is significantly lower than for Patient 1 due to the increased spatial data coverage.

The validation scores for LAT prediction for the virtual patient show that ISE coverage ($|\text{ISE}| \leq 2$) is consistently $\sim 95\%$, and the nRMSE is already less than $1.0\%$ for $250$ observations. Note that these scores were comparing all LAT predictions against all ground truth LAT values. These scores demonstrate excellent prediction of LAT using the GPMI method.

\subsection{Conduction velocity}

\begin{figure}[h]
	\centering
    \caption{CV vectors at face centroids colored by CV magnitude. CV from simulation (top), and CV predictions using 1000 observations randomly selected from the simulation: noiseless observations (middle) and noisy observations (bottom).} \label{fig:CV-sim}
       \includegraphics[width=0.50\textwidth]{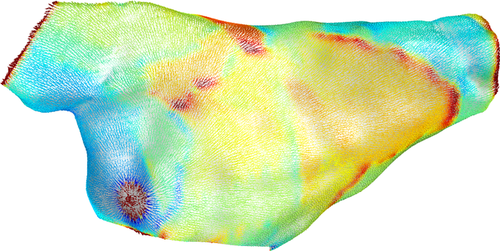}
       \includegraphics[width=0.450\textwidth]{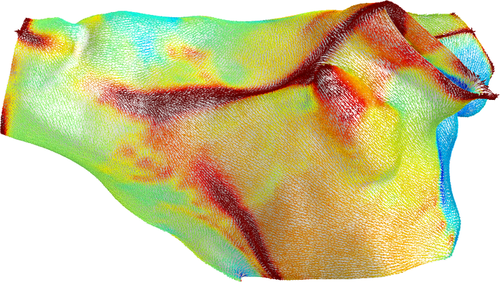}
    %
    %
       \includegraphics[width=0.50\textwidth]{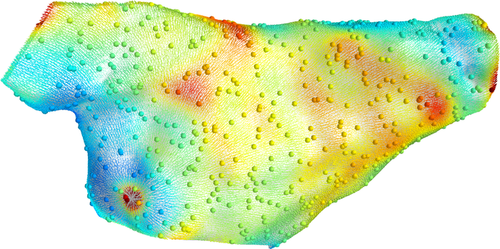}
        \includegraphics[width=0.450\textwidth]{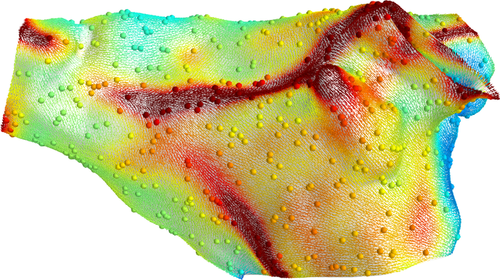}
    %
	%
    %
       \includegraphics[width=0.50\textwidth]{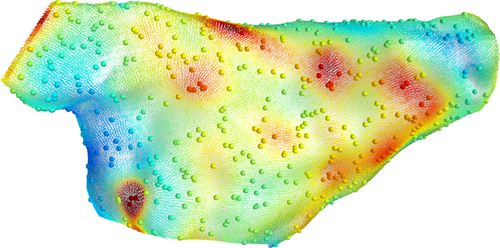}
        \includegraphics[width=0.450\textwidth]{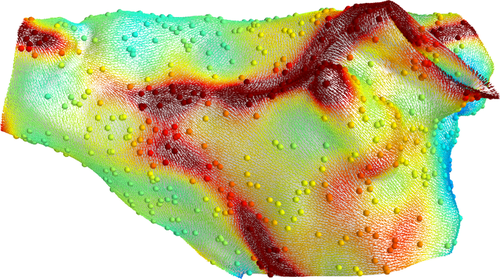}
    \\ \vspace{0.20cm}
       \includegraphics[width=0.475\textwidth]{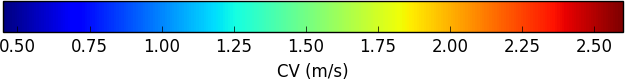}
	\end{figure}

Figure \ref{fig:CV-sim} shows CV for the virtual patient. In the top row, CV is calculated at centroids using the 'element' method equation \eqref{eq:CVcalc} with known (ground truth) LAT at the mesh vertices. The second and third rows show CV prediction using 1000 random observations of LAT at mesh vertices, with random noise $\epsilon \sim \mathcal{N}(0,1)$ corrupting these observations for the third row. The predictions capture the heterogeneous CV field produced in the simulation, though the features are more smoothed in the prediction, especially where the true CV varied on a lengthscale shorter than the resolution given by the distances between observations. The predicted CV field is slightly smoother in the case with noisy LAT observations. Overall, the predictions seem to perform very well, capturing even the locations of wave collision and stimulus location.

\begin{figure}[h]
	\centering
    \caption{CV vectors at face centroids colored by CV magnitude interquartile range, for the predictions in Figure \ref{fig:CV-sim}. Note that the color scales are different for each case, due to differences in prediction uncertainty for CV magnitude.} \label{fig:CV-sim-uq}
	%
	%
       \includegraphics[width=0.50\textwidth]{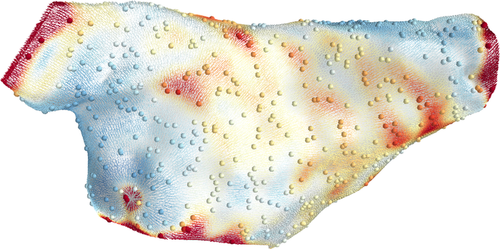}
       \includegraphics[width=0.450\textwidth]{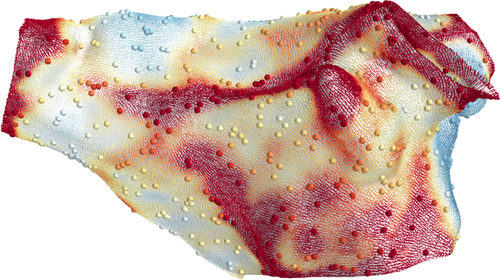}
    \\ \vspace{0.20cm}
       \includegraphics[width=0.50\textwidth]{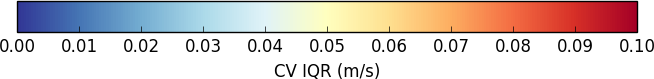}
	\\
	%
	%
       \includegraphics[width=0.50\textwidth]{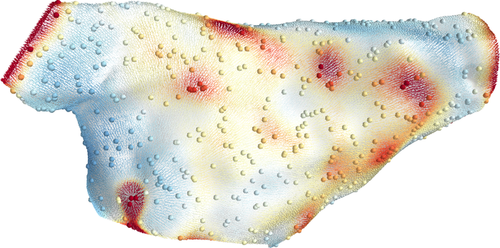}
       \includegraphics[width=0.450\textwidth]{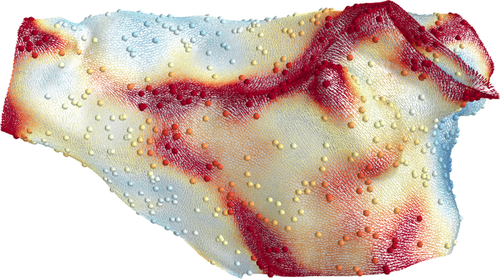}
    \\ \vspace{0.20cm}
       \includegraphics[width=0.50\textwidth]{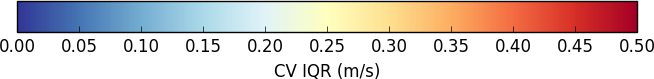}
\end{figure}

Figure \ref{fig:CV-sim-uq} shows the prediction uncertainty (interquartile range) for CV magnitude, corresponding to the predictions in Figure \ref{fig:CV-sim}. There is significantly more uncertainty in CV for prediction using noisy LAT observations (the colorbar scales are different). Higher CV correlates with higher prediction uncertainty (as is also the trend for Patient 2, discussed below).

Table \ref{tab:val} gives the validation scores for the magnitude of the LAT gradient for the virtual patient. The scores are very similar for both methods of calculating the ground truth. Notably, the nRMSE scores are an order of magnitude larger for $|\nabla\text{LAT}|$ prediction than for LAT prediction, although the ISE scores suggest that the truth is captured well by the uncertainty. For $250$ observations, the nRMSE score for $|\nabla\text{LAT}|$ is less than $10\%$ (and the nRMSE score for LAT is less than $1\%$. We propose that $250$ observations (spaced over the atria) could be considered a lower bound for having confidence in CV predictions calculated with our methodology.

\begin{figure}[h]
	\centering
    \caption{Validation plots for the magnitude of the gradient of LAT. Simulation (ground truth) values were obtained with the mesh flattening method for 1000 well spaced centroid locations. For prediction, 1000 noisy LAT observations were used. The error bars on the prediction vs truth plot (left) represent the $9$th and $91$st percentiles.} \label{fig:compare}
       \includegraphics[width=0.475\textwidth]{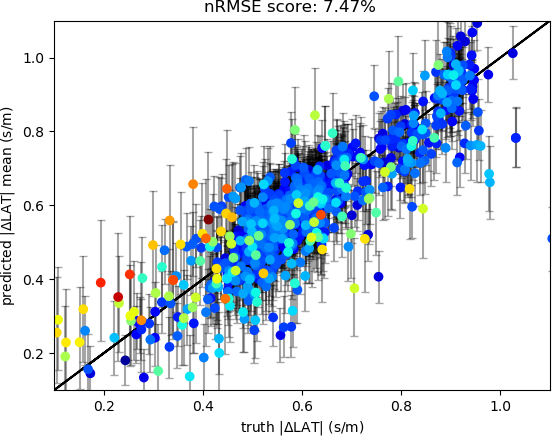}
       \includegraphics[width=0.475\textwidth]{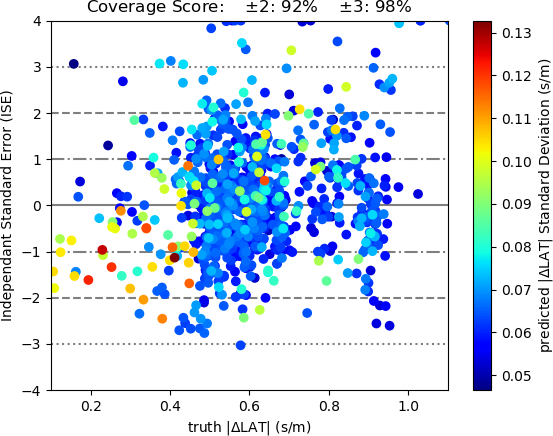}
	\end{figure}

Figure \ref{fig:compare} shows prediction vs observation plots (left) and ISE plots (right) for the virtual patient for $1000$ noisy observations, where the ground truth is for $|\nabla\text{LAT}|$ calculated by the 'wave'. Points are ordered so that larger standard deviations are plotted in the foreground. The error bars in the prediction vs observation plots correspond to the $9$th and $91$st percentiles. The uncertainty mostly captures the ground truth. Notably, the uncertainty is highest where $|\nabla\text{LAT}|$ is overestimated (i.e. CV is underestimated) relative to the ground truth. This is likely because regions of fast CV cannot be captured when uncertainty is high.

\begin{figure}[t]
	\centering
    \caption{CV predictions for Patient 1, shown as CV vectors at face centroids colored by CV magnitude (top) and interquartile range (bottom). Spheres show these quantities at observation vertices (interpolated from neighbouring faces).} \label{fig:CV-S1}
       \includegraphics[width=0.450\textwidth]{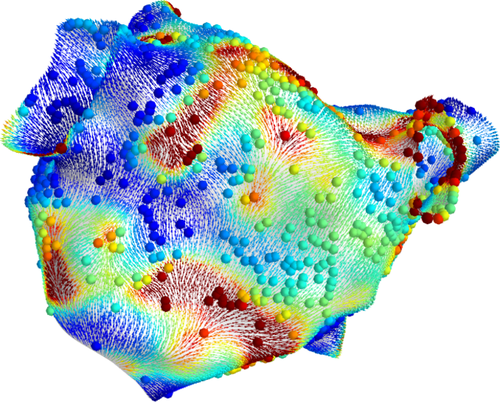}
       \includegraphics[width=0.450\textwidth]{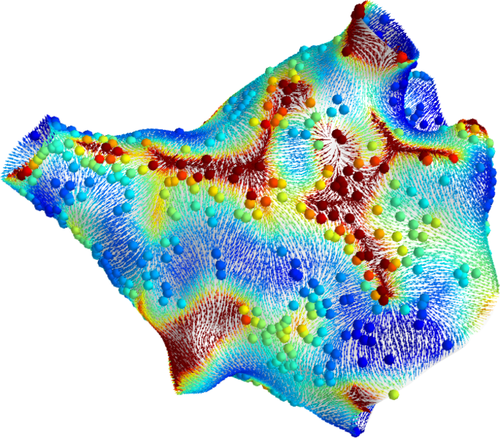}
    \\ \vspace{0.20cm}
       \includegraphics[width=0.50\textwidth]{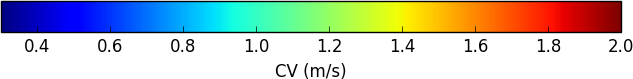}
    \\
       \includegraphics[width=0.450\textwidth]{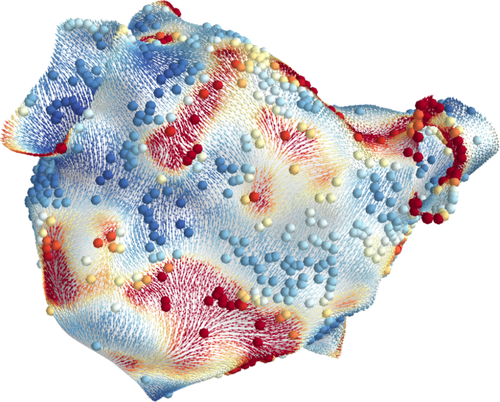}
       \includegraphics[width=0.450\textwidth]{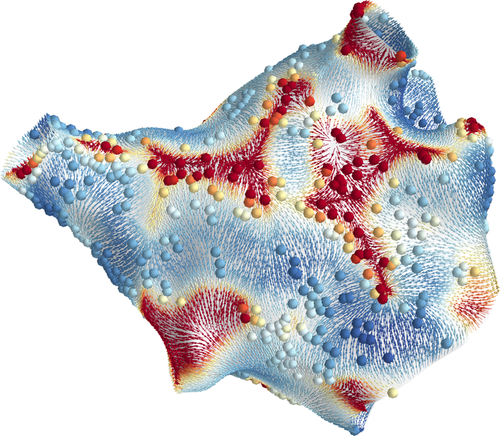}
    \\ \vspace{0.20cm}
       \includegraphics[width=0.50\textwidth]{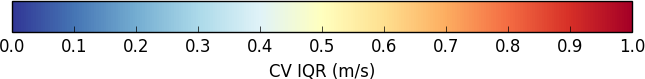}
\end{figure}

For Patient 1, there are only $61$ LAT observations, much less than the 250 minimal number of observations indicated from the simulation studies. These observations are also highly clustered such that large sections of the mesh have no observations at all (see Figure \ref{fig:LAT-apsaf}). Therefore, we do not present plots of CV magnitude for Patient 1. Figure \ref{fig:CV-S1} shows CV calculated from the posterior of the LAT gradient for Patient 2, coloured by the CV magnitude (top) and the interquartile range (bottom). The colored spheres represent these values (interpolated from neighbouring faces) at observation vertices. The predicted CV magnitude is sometimes very high far from the observations, where the GP flattens out beyond the data (shallow LAT gradients correspond to high CV), and in regions of wave collision where LAT is interpolated across two different wavefronts (the calculated value of CV does not represent wave propagation).

CV uncertainty depends highly on distance from observations, but also on the direction of the (predicted) CV field, so that physiological and more certain CV predictions are made where there are several nearby observations aligned along the direction of (predicted) wave propagation. This makes physical sense as observations aligned along the direction of changing LAT are required to inform the gradient of LAT, and is especially useful since we do not know the direction of propagation in advance of the interpolation.

\section{Discussion} \label{sec:Discussion}

In this study we used Gaussian processes to interpolate LAT directly on atrial manifolds in order to calculate CV distributions. The results show that point-wise uncertainty in LAT interpolation is not sufficient for inferring uncertainty in CV. Rather, CV uncertainty depends on LAT observation uncertainty, data density, and the presence of sufficient observations along the direction of wave propagation. Also, larger CV will generally have higher associated uncertainties due to the inverse relationship between the gradient of LAT and CV, such that the same uncertainty in LAT will have much larger effects on CV for shallow LAT gradients (for a given spacing of percentiles for the magnitude of the gradient of LAT, the percentiles on the distribution for CV will become further apart as the gradient of LAT is decreased). Precisely measuring fast CV is therefore difficult. Uncertainty in CV restitution curves could be important for calibrating atrial models, e.g., in \cite{Corrado2017, Corrado2018} where the CV uncertainty could weight the comparison between simulated and observed restitution curves.


We have focused here on solving the geometry aspect of the problem of probabilistic interpolation on atrial manifolds, which had only been possible previously for the special case of Gaussian Markov random fields (as in our previous publication \cite{Coveney2019}). The method presented here, Gaussian Process Manifold Interpolation (GPMI), allows us to solve the problem of calculating CV statistics from a probabilistic LAT interpolation. It would also be suitable for probabilistic interpolation of other quantities such as voltage and tissue parameters. The flexible modeling framework presented here means that \emph{observations} of LAT gradient could also be included into the modeling. This could allow both LAT and LAT gradient observations (for example, from multi-polar catheters where consideration of wave propagation is used to obtain precise CV values \cite{Roney2019, Verma2018}), to be used in a unified and consistent modelling framework. It may also be possible to include linear constraints on interpolation via the covariance function \cite{Jidling2017}, and to directly include physics using latent force models as in \cite{Alvarado2014}.

In theory, it should be possible to apply this modelling framework to a 3D domain such as the ventricles. Besides the increase in the number of basis functions that may be required to accurately expand the kernel \cite{Solin2019}, the main difficult may be in the boundary conditions at the \emph{surface} of the domain. Some form of mesh extension analogous to that done here may be possible, or perhaps different boundary conditions might be applied. One possibility may be to utilize a separable kernel structure, similar to the 'spatial-pacing' model mentioned in Section \ref{sec:methods}\ref{sec:LATwithGP}, whereby the 'thickness' direction is accounted for with a separate kernel that multiplies the manifold kernel.

\section{Limitations}

LAT interpolation is difficult in part because of genuine discontinuities where wavefronts collide and where lines of block may exist, and also because of the quasi-monotonic nature of the underlying activation field. These difficulties are not solved with a Gaussian process, which may further introduce artefacts such as reverting to the mean in regions of extrapolation and oscillations in larger gaps between observations. Measurements of CV made by fitting plane waves to LAT observations can explicitly model a propagating wavefront, but the LAT interpolation here does not. Furthermore, CV calculations using LATs measured from individual placements of a multipolar catheter have the advantage that the distance between recording sites is fixed and known, whereas LAT interpolation and CV prediction from combining observations for many separate placements of a catheter (as done here) is potentially subject to large inaccuracies, i.e., some non-physiological CV predictions simply reflect the data, and it may be hard to correct for this without making strong prior assumptions about CV.

Accurately predicting  the gradient of LAT is more challenging than predicting LAT, as noisy observations of LAT provide limited gradient information, particularly when sparsely spaced. Furthermore, the \emph{magnitude} of the LAT gradient is not a Gaussian process. Nonetheless, we find that the calibration of the predictions (the coverage achieved by credible intervals averaged over all predictions) is reasonable for the posterior distribution of the LAT gradient magnitude.

\section{Conclusion}

We have introduced the Gaussian Process Manifold Interpolation (GPMI) method which allows us to perform probabilistic interpolation directly on atrial manifolds using a reduced-rank Gaussian process that expresses the covariance function in terms of eigenfunctions of the Laplacian operator on a mesh triangulation of the manifold. 
This allows us to interpolate uncertain local activation times with a differentiable model, which then allows us to calculate probabilistic estimates of conduction velocity maps. In general, this method allows Gaussian processes to be used to model data on irregular spatial domains such as the left atrium, while directly accounting for the topology of the domain and the physically meaningful distances that correlations ought to depend upon.

\vskip6pt

\enlargethispage{20pt}

Ethics: Clinical data were recorded as part of routine care for first-time ablation for atrial fibrillation. Ethical approval was granted by the National Research Ethics Service (10/H0802/77) for Patient 1, and (REC number 18/HRA/0083) for Patient 2.


Contributions: SC, RW, JO and RC conceived the study. SC designed and implemented the main methods. CC, CHR and DOH processed the raw clinical data. SW and MN performed the clinical data collection. All authors read and approved the manuscript.

Competing interests: The authors declare that they have no competing interests.

Funding: This work was funded by grants from the UK Engineering and Physical Sciences Research Council (EP/M012492/1, NS/A000049/1, EP/P01268X/1, EP/K037145/1), the British Heart Foundation (PG/15/91/31812, PG/13/37/30280), Kings Health Partners London National Institute for Health Research (NIHR) Biomedical Research Centre, the Wellcome/EPSRC Centre for Medical Engineering [WT 203148/Z/16/Z], Medical Research Council Skills Development Fellowship (MR/S015086/1), and funding for research from Abbott, EBR systems, Pfizer, and Siemens.

Acknowledgements: We are grateful to Dr Mauricio Alvarez, Dr Arno Solin, and Prof Finn Lindgren for helpful discussions.



\bibliographystyle{ieeetr}
\bibliography{references}

\begin{thebibliography}{10}

\bibitem{Cantwell2015}
C.~D. Cantwell, C.~H. Roney, F.~S. Ng, J.~H. Siggers, S.~J. Sherwin, and N.~S.
  Peters, ``Techniques for automated local activation time annotation and
  conduction velocity estimation in cardiac mapping,'' {\em Computers in
  Biology and Medicine}, 2015.

\bibitem{Honarbakhsh2019}
S.~Honarbakhsh, R.~J. Schilling, M.~Orini, R.~Providencia, M.~Finlay,
  E.~Keating, P.~D. Lambiase, A.~Chow, M.~J. Earley, S.~Sporton, and R.~J.
  Hunter, ``{Left atrial scarring and conduction velocity dynamics: Rate
  dependent conduction slowing predicts sites of localized reentrant atrial
  tachycardias},'' {\em International Journal of Cardiology}, vol.~278,
  pp.~114--119, 2019.

\bibitem{Bellmann2019}
B.~Bellmann, M.~Zettwitz, T.~Lin, P.~Ruppersberg, S.~Guttmann, V.~Tscholl,
  P.~Nagel, M.~Roser, U.~Landmesser, and A.~Rillig, ``{Velocity characteristics
  of atrial fibrillation sources determined by electrographic flow mapping
  before and after catheter ablation},'' {\em International Journal of
  Cardiology}, vol.~286, pp.~56--60, 2019.

\bibitem{Corrado2018}
C.~Corrado, S.~Williams, R.~Karim, G.~Plank, M.~O'Neill, and S.~Niederer, ``{A
  work flow to build and validate patient specific left atrium
  electrophysiology models from catheter measurements},'' {\em Medical Image
  Analysis}, vol.~47, pp.~153--163, 2018.

\bibitem{Boyle2019}
P.~M. Boyle, T.~Zghaib, S.~Zahid, R.~L. Ali, D.~Deng, W.~H. Franceschi, J.~B.
  Hakim, M.~J. Murphy, A.~Prakosa, S.~L. Zimmerman, H.~Ashikaga, J.~E. Marine,
  A.~Kolandaivelu, S.~Nazarian, D.~D. Spragg, H.~Calkins, and N.~A. Trayanova,
  ``Computationally guided personalized targeted ablation of persistent atrial
  fibrillation,'' {\em Nature Biomedical Engineering}, vol.~3, pp.~870–--879,
  2019.

\bibitem{Verma2018}
B.~Verma, T.~Oesterlein, A.~Loewe, A.~Luik, C.~Schmitt, and O.~D{\"o}ssel,
  ``Regional conduction velocity calculation from clinical multichannel
  electrograms in human atria,'' {\em Computer in Biology and Medicine},
  vol.~92, pp.~188--196, 2018.

\bibitem{Loewe2019}
A.~Loewe, E.~Poremba, T.~Oesterlein, A.~Luik, C.~Schmitt, G.~Seemann, and
  O.~D{\"o}ssel, ``Patient-specific identification of atrial flutter
  vulnerability–a computational approach to reveal latent reentry pathways,''
  {\em Frontiers in Physiology}, vol.~9, p.~1910, 2019.

\bibitem{Roney2019}
C.~H. Roney, J.~Whitaker, I.~Sim, L.~O'Neill, R.~K. Mukherjee, O.~Razeghi,
  E.~J. Vigmond, M.~Wright, M.~D. O'Neill, S.~E. Williams, and S.~A. Niederer,
  ``{A technique for measuring anisotropy in atrial conduction to estimate
  conduction velocity and atrial fibre direction},'' {\em Computers in Biology
  and Medicine}, vol.~104, no.~October 2018, pp.~278--290, 2019.

\bibitem{Coveney2019}
S.~Coveney, C.~Corrado, C.~Roney, R.~Wilkinson, J.~Oakley, F.~Lindgren,
  S.~Williams, M.~D. O'Neill, S.~Niederer, and R.~H. Clayton, ``{Probabilistic
  Interpolation of Uncertain Local Activation Times on Human Atrial
  Manifolds},'' {\em IEEE Transactions on Biomedical Engineering}, vol.~67,
  pp.~99--109, 2019.

\bibitem{Lindgren2011}
F.~Lindgren, H.~Rue, and J.~Lindstr{\"{o}}m, ``{An explicit link between
  Gaussian fields and Gaussian Markov random fields: the stochastic partial
  differential equation approach},'' {\em Journal of the Royal Statistical
  Society, Series B}, vol.~73, pp.~423--498, 2011.

\bibitem{Rasmussen2006}
C.~E. Rasmussen, C.~K.~I. Williams, R.~S. Sutton, A.~G. Barto, P.~Spirtes,
  C.~Glymour, R.~Scheines, B.~Sch{\"{o}}lkopf, and A.~J. Smola, ``{Gaussian
  Processes for Machine Learning},'' 2006.

\bibitem{Solin2019}
A.~Solin and S.~S{\"{a}}rkk{\"{a}}, ``{Hilbert space methods for reduced-rank
  Gaussian process regression},'' {\em Statistics and Computing}, 2019.

\bibitem{Wood1995}
A.~T. Wood, ``When is a truncated covariance function on the line a covariance
  function on the circle?,'' {\em Statistics \& probability letters}, vol.~24,
  no.~2, pp.~157--164, 1995.

\bibitem{Kok}
A.~Solin and M.~Kok, ``{Know Your Boundaries : Constraining Gaussian Processes
  by Variational Harmonic Features},'' {\em arXiv}, p.~1904.05207v1, 2019.

\bibitem{Sorkine2005}
O.~Sorkine, ``{Laplacian Mesh Processing},'' {\em Eurographics - State of the
  Art Reports}, no.~Section 4, pp.~53--70, 2005.

\bibitem{Reuter2009}
M.~Reuter, S.~Biasotti, D.~Giorgi, G.~Patan{\`{e}}, and M.~Spagnuolo,
  ``{Discrete Laplace-Beltrami operators for shape analysis and
  segmentation},'' {\em Computers and Graphics (Pergamon)}, vol.~33, no.~3,
  pp.~381--390, 2009.

\bibitem{SciPy}
P.~{Virtanen}, R.~{Gommers}, T.~E. {Oliphant}, M.~{Haberland}, T.~{Reddy},
  D.~{Cournapeau}, E.~{Burovski}, P.~{Peterson}, W.~{Weckesser}, J.~{Bright},
  S.~J. {van der Walt}, M.~{Brett}, J.~{Wilson}, K.~{Jarrod Millman},
  N.~{Mayorov}, A.~R.~J. {Nelson}, E.~{Jones}, R.~{Kern}, E.~{Larson},
  C.~{Carey}, {\.I}.~{Polat}, Y.~{Feng}, E.~W. {Moore}, J.~{Vand erPlas},
  D.~{Laxalde}, J.~{Perktold}, R.~{Cimrman}, I.~{Henriksen}, E.~A. {Quintero},
  C.~R. {Harris}, A.~M. {Archibald}, A.~H. {Ribeiro}, F.~{Pedregosa}, P.~{van
  Mulbregt}, and S.~.~. {Contributors}, ``{SciPy 1.0--Fundamental Algorithms
  for Scientific Computing in Python},'' {\em arXiv e-prints},
  p.~arXiv:1907.10121, Jul 2019.

\bibitem{quLATi}
S.~Coveney, ``{quLATi}.'' {https://zenodo.org/record/3758043\#.Xp2s5nVKjeQ}.

\bibitem{Vigmond2003}
E.~J. Vigmond, M.~Hughes, G.~Plank, and L.~J. Leon, ``Computational tools for
  modeling electrical activity in cardiac tissue,'' {\em J Electrocardiol},
  vol.~36 Suppl, pp.~69--74, 2003.

\bibitem{Corrado2017}
C.~Corrado, J.~Whitaker, H.~Chubb, S.~E. Williams, M.~Wright, J.~Gill, M.~D.
  ONeill, S.~A. Niederer, M.~O'Neill, and S.~Niederer, ``{Personalized Models
  of Human Atrial Electrophysiology Derived From Endocardial Electrograms},''
  {\em IEEE Transactions on Biomedical Engineering}, vol.~64, no.~4,
  pp.~735--742, 2017.

\bibitem{Jidling2017}
C.~Jidling, N.~Wahlstr{\"{o}}m, A.~Wills, and T.~B. Sch{\"{o}}n, ``{Linearly
  constrained Gaussian processes},'' no.~Nips, 2017.

\bibitem{Alvarado2014}
P.~A. Alvarado, M.~A. Alvarez, G.~Daza-Santacoloma, A.~Orozco, and
  G.~Castellanos-Dominguez, ``{A latent force model for describing electric
  propagation in deep brain stimulation: A simulation study},'' {\em 2014 36th
  Annual International Conference of the IEEE Engineering in Medicine and
  Biology Society, EMBC 2014}, pp.~2617--2620, 2014.

\end{thebibliography}




\end{document}



\title{Gaussian Process Manifold Interpolation for Probabilistic Atrial Activation Maps and Uncertain Conduction Velocity\\ \Large \bf Supplementary Material}


\author{
Sam~Coveney$^{1}$, Cesare~Corrado$^{2}$, Caroline~H~Roney$^{2}$,\\
Daniel~O'Hare$^{2}$, Steven~E~Williams$^{2}$, Mark~D O'Neill$^{2}$,\\ Steven~A~Niederer$^{2}$, Richard~H~Clayton$^{1}$,\\
Jeremy~E~Oakley$^{3}$, and Richard~D~Wilkinson$^{3}$}


\maketitle

\begin{flushleft}{
\small
\bf{1}: Insigneo Institute for \emph{in-silico} Medicine, University of Sheffield, Sheffield, UK.\\
\bf{2}: Division of Imaging Sciences and Biomedical Engineering, King's College London, UK.\\
\bf{3}: School of Mathematics and Statistics, University of Sheffield, UK.\\
$^{\ast}$ E-mail: s.coveney@sheffield.ac.uk
}\end{flushleft}


\section{Data collection}

\textbf{Patient 1:} Data were obtained as part of a study involving patients with paroxysmal atrial fibrillation and undergoing first-time atrial fibrillation ablation. Ethical approval was granted by the National Research Ethics Service (10/H0802/77). Following femoral access and trans-septal puncture, two 8.5 French SR0 long sheaths and a PentaRay mapping catheter (Biosense Webster, CA, 1mm electrode size, 4-4-4mm spacing) were advanced into the left atrium. Decapole (St Jude Medical, MN) and pentapole (Bard Electrophysiology, MA) catheters were positioned in the coronary sinus (CS) and high right atrium (HRA), respectively. An S1-S2 programmed pacing protocol was applied, consisting of a 2-beat drivetrain with a cycle length S1=470~ms followed by a single premature extra stimulus S2$<$S1. The S1-S2 interval was reduced continuously and without operator interference from 343~ms to 200~ms (or loss of capture) by reducing the current S1-S2 interval by 2\% of its value (rounded to nearest ms). All pacing stimuli were delivered at a voltage of at least twice threshold and with a pulse width of 2~ms, from either the CS or HRA. Bipolar electrograms (EGMs) were recorded at a sampling rate of 4~kHz using the 10 electrodes of the PentaRay catheter for each S1-S2 interval. The catheter was then moved to another location on the left atrial endocardium, and the process repeated for up to 15 locations. For each S1-S2 interval, we discarded any EGM traces not containing a discernible activation complex. We obtained left atrial anatomies during the clinical procedures using an electro-anatomical mapping system (St Jude Velocity). 

\bigskip
\noindent\textbf{Patient 2:} Data were collected as part of routine clinical care for first-time pulmonary vein isolation to treat persistent atrial
fibrillation. Ethical approval was granted
for retrospective data analysis (REC number 18/HRA/0083). Prior to radiofrequency ablation, left atrial geometry and a paced map were obtained using an Electroanatomical mapping system (CARTO3, Biosense Webster, USA) during constant pacing at a cycle length of 500ms from the proximal (9,10) dipoles of a decapolar catheter positioned in the coronary sinus (CS). Endocardial contact points were acquired using a circular mapping catheter (Lasso 2515; Biosense Webster, USA) introduced to the left atrium via transseptal access.

\section{Mesh extension}

\begin{figure}[!htbp]
    \centering
      \includegraphics[scale=0.5]{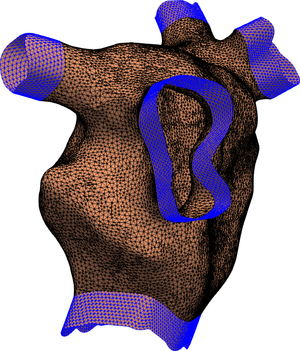}
    \caption{Mesh extension  for Patient 2.} \label{fig:meshExtend}
\end{figure}

In order to remove the effects of the Laplacian boundary conditions from the interpolation on the atrial mesh, we extend the mesh with elements as in Figure \ref{fig:meshExtend}. This does not change the mesh topology, and the shortest distance between points on the original mesh is almost completely unchanged (with small exceptions caused by non-planer clipping of the clinical mesh to produce the holes in the interpolation mesh; for planar or convex edges, the shortest distances would be completely unchanged).

The algorithm is as follows. For each hole in the mesh (e.g. mitral valve), the vertices belonging to the `edge' describing this hole are identified (face edges at the mesh edge belong to one face only; this can be used to identify vertices on the mesh edges). These vertex coordinates are used to fit a plane, and the plane normal $\vec{\mathbf{n}}$ sign is adjusted such that $\vec{\mathbf{n}}$ points away from the mesh centre of mass. Vector $\vec{\mathbf{n}}$ points in the direction that layers of elements will be added. Then, starting on any vertex in the edge and proceeding either clockwise or anti-clockwise until all edge vertices have been considered, the following is done:

\begin{enumerate}
    \item calculate the centre $\mathbf{x_p}$ between current vertex $\mathbf{x_i}$ and neighbouring vertex $\mathbf{x_j}$ 
    \item add a new vertex at position $\mathbf{x_n} = \mathbf{x_p} +  \vec{\mathbf{n}} \times |\mathbf{x_j} - \mathbf{x_i}|  \tan(60\deg) / 2$
    \item append this new vertex $\mathbf{x_n}$ to the list of all vertices 
    \item append a new element connecting $\mathbf{x_i}, \mathbf{x_j}, \mathbf{x_n}$ to the list of all elements
\end{enumerate}

Now the edge will look like the teeth of a saw. New elements must be added to the list of all elements that connect the newly created vertices to each other, and to the original vertex nearest both of them. This concludes the creation of a single layer of new elements. This process can be repeated to build up sufficient layers to prevent the boundary conditions from producing artefacts in the interpolation (the plane normal vectors can be recalculated for each layer, or the original vector used for all layers).

\section{Explained Variance}

\begin{figure}[h]
	\centering
    \caption{Variance captured by including the first $m$ eigenfunctions.} \label{fig:screeplot}
    \includegraphics[width=0.750\textwidth]{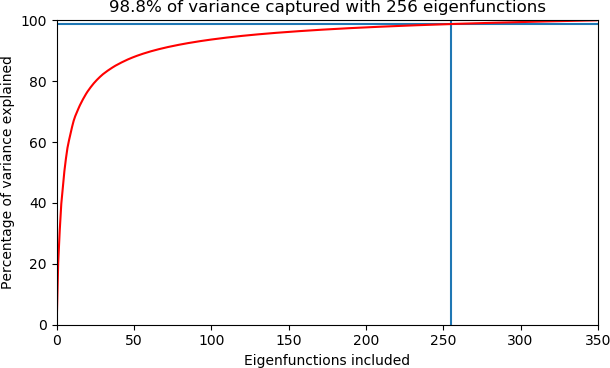}
\end{figure}

The percentage of variance captured by the first $m$ basis functions can be calculated as
\begin{equation}
    \text{EV}_m = 100 \times \sum_{j=1}^{m} S(\sqrt{\lambda_j}) \bigg/ \sum_{j=1}^{M} S(\sqrt{\lambda_j}) \quad \quad \text{where} \quad \quad 1 \leq m \leq M
\end{equation}
where $S$ is the kernel spectral density evaluated at the square root of the eigenvalue $\lambda_j$ (kernel hyperparameters have been optimized). Figure \ref{fig:screeplot} shows, for the model fit of the virtual patient with $1000$ noisy observations and $M = 350$ basis functions, that the first $256$ eigenfunctions capture $98.8\%$ of the variance. Therefore $256$ eigenfunctions are more than sufficient for LAT interpolation.

\section{Simulation Parameters}

\begin{figure}[h]
	\centering
    \caption{Parameter values for the simulation (virtual patient).} \label{fig:params}
       \includegraphics[width=0.490\textwidth]{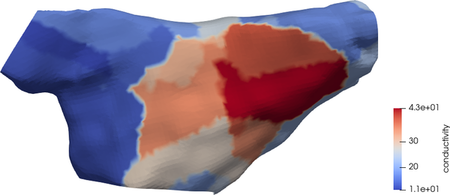}
       \includegraphics[width=0.490\textwidth]{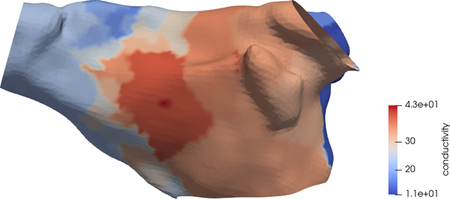}
    \\ \vspace{0.20cm}
    %
       \includegraphics[width=0.490\textwidth]{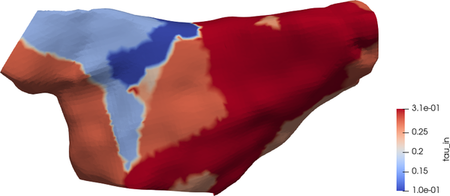}
       \includegraphics[width=0.490\textwidth]{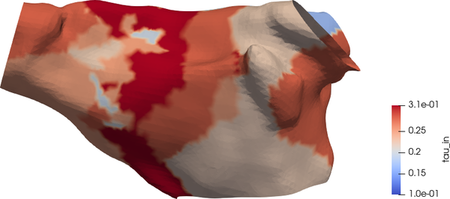}
	\\ \vspace{0.20cm}
    %
       \includegraphics[width=0.490\textwidth]{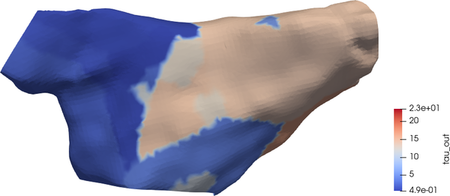}
       \includegraphics[width=0.490\textwidth]{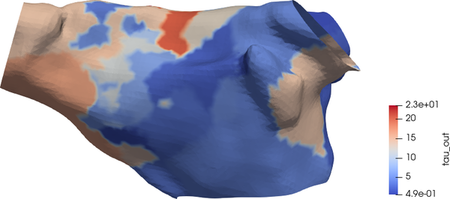}
	\\ \vspace{0.20cm}
        %
       \includegraphics[width=0.490\textwidth]{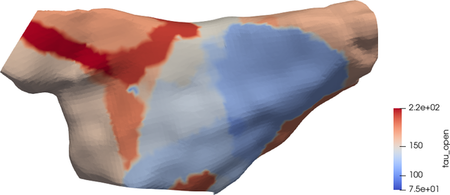}
       \includegraphics[width=0.490\textwidth]{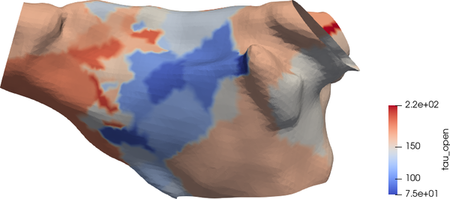}
	\\ \vspace{0.20cm}
    %
       \includegraphics[width=0.489  \textwidth]{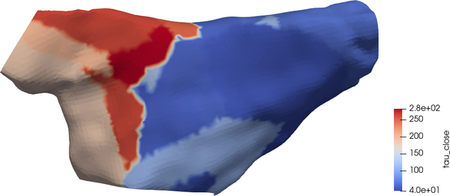}
       \includegraphics[width=0.489\textwidth]{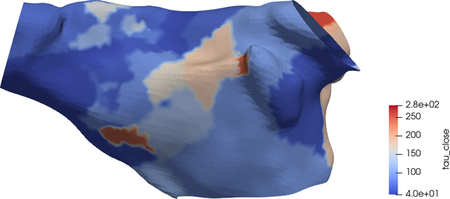}
\end{figure}

Figure \ref{fig:params} shows the parameter values used for simulation of the virtual patient. The parameter values are down-sampled to nodes for visualization purposes. The $\tau$ parameter units are milliseconds, and the conductivity units are in $\text{cm}^2/\text{s}$. The CARP simulation files can be found at \cite{zenodo-dataset}.

